\documentclass{article}

\usepackage{microtype}
\usepackage{graphicx}
\usepackage{subfigure}
\usepackage{booktabs} 

\usepackage{hyperref}

\usepackage{multibib}
\newcites{Apx}{Reference}
\usepackage{multirow}
\usepackage{graphicx}
\usepackage{wrapfig}
\usepackage{amsmath,bm,bbm,upgreek}
\usepackage{amsthm}
\usepackage{amssymb}
\usepackage{pifont}
\usepackage{xcolor}
\usepackage{caption}
\usepackage{enumitem}
\usepackage{subcaption}
\usepackage{url}
\usepackage{comment}

\usepackage[accepted]{icml2024}

\usepackage{amsmath}
\usepackage{amssymb}
\usepackage{mathtools}
\usepackage{amsthm}

\usepackage[capitalize,noabbrev]{cleveref}

\theoremstyle{plain}
\newtheorem{theorem}{Theorem}
\newtheorem{lemma}{Lemma}

\newtheorem{manuallemmainner}{Lemma}
\newenvironment{manuallemma}[1]{%
  \renewcommand\themanuallemmainner{#1}%
  \manuallemmainner
}{\endmanuallemmainner}

\newtheorem{manualtheoreminner}{Theorem}
\newenvironment{manualtheorem}[1]{%
  \renewcommand\themanualtheoreminner{#1}%
  \manualtheoreminner
}{\endmanualtheoreminner}

\usepackage[textsize=tiny]{todonotes}

\icmltitlerunning{Dynamic Latent Separation for Deep Learning}

\begin{document}

\twocolumn[
\icmltitle{Dynamic Latent Separation for Deep Learning}



\icmlsetsymbol{equal}{*}

\begin{icmlauthorlist}

\icmlauthor{Yi-Lin Tuan}{ucsb}
\icmlauthor{Zih-Yun Chiu}{ucsd}
\icmlauthor{William Yang Wang}{ucsb}
\end{icmlauthorlist}

\icmlaffiliation{ucsb}{Department of Computer Science, University of California Santa Barbara}
\icmlaffiliation{ucsd}{Department of Electrical and Computer Engineering, University of California San Diego}

\icmlcorrespondingauthor{Yi-Lin Tuan}{ytuan@cs.ucsb.edu}

\icmlkeywords{Machine Learning, ICML}

\vskip 0.3in
]



\printAffiliationsAndNotice{}  

    \begin{abstract}
        
A core problem in machine learning is to learn expressive latent variables for model prediction on complex data that involves multiple sub-components in a flexible and interpretable fashion.
Here, we develop an approach that improves expressiveness, provides partial interpretation, and is not restricted to specific applications.
The key idea is to dynamically distance data samples in the latent space and thus enhance the output diversity.
Our dynamic latent separation method, inspired by atomic physics, relies on the jointly learned structures of each data sample, which also reveal the importance of each sub-component for distinguishing data samples.
This approach, {\it atom modeling}, requires no supervision of the latent space and allows us to learn extra partially interpretable representations besides the original goal of a model. 
We empirically demonstrate that the algorithm also enhances the performance of small to larger-scale models in various classification and generation problems.
    \end{abstract}

    \section{Introduction}

Deep neural networks with multiple hidden layers are trained to be expressive models that learn complicated relationships between their inputs and outputs~\cite{srivastava2014dropout}. 
Among various data types, data samples that consist of many sub-units, such as images and texts, can require models to be more expressive to consider nuanced differences among sub-units.
The demand for this delicacy leads to developing large-scale and complex model architectures~\cite{vaswani2017attention}, which cause drawbacks such as compromised model interpretability~\cite{ribeiro2016should, bastani2017interpreting, rudin2019stop, jain2019attention}.

Various algorithms exist that improve model expressiveness not by advancing model architectures. 
For instance, contrastive learning ameliorates classification expressiveness~\cite{dosovitskiy2014discriminative,chen2020simple} by pushing away latent features from different classes. 
Vector quantization tackles the expressiveness of autoencoders~\cite{van2017neural} by learning discrete representations using a preset codebook.
As a separate effort, post-hoc methods or designed models that follow self-explaining protocol~\cite{alvarez2018towards} reveal some underlying reason for model behaviors. 
While these methods show promising results in their bundled applications, it is yet certain of their transferability and usefulness to other applications.
Meanwhile, it is yet underexplored of generalizable training algorithms that can simultaneously help expressiveness and uncover partial explanations.

\begin{figure*}[t]
    \centering
    \includegraphics[width=.95\linewidth]{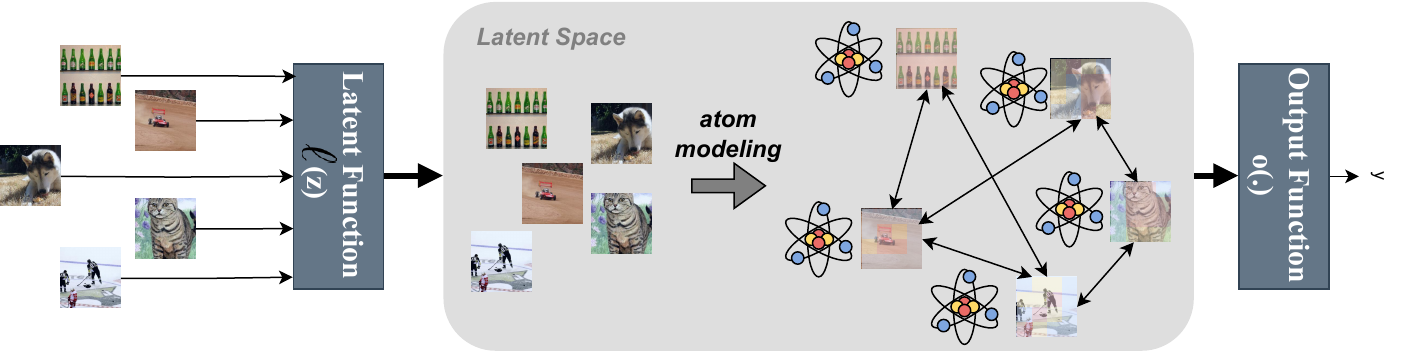}
    \caption{Illustration of atom modeling use case. Consider a model $f_\theta = o(\ell(\mathbf{z}))$; data samples are transformed into the latent space and their latent representations are distanced using atom modeling associated with the training criterion for output $y$. The colors labeled on each image in the latent space present the learned {\it token importance} that indicates which part is more crucial to identify data samples.}
    \label{fig:atom}
\end{figure*}

We present a novel algorithm that simultaneously improves model expressiveness, provides an interpretation of sub-component importance, and is generalizable to multiple applications. 
Our method, {\it atom modeling}, first maps the latent representations of each sub-component in a data sample to a learnable {\it token importance} and then {\it dynamically distances data samples} based on token importance using a loss function inspired by Coulomb force~\cite{coulomb1785premier}. 
After training, token importance reveals which sub-components in a data sample contribute to its semantic meaning and are key to distinguishing itself from other data samples. 
The dynamic separation between data samples encourages a model to predict diverse outputs, thus boosting expressiveness.

This method can be viewed as connecting sub-component importance and inter-sample relationships to elevate impacts from local details.
A similar observation can be found in atomic physics, where the balance distance between atoms, fundamental particles that form every matter in nature, depends on the structure of sub-atomic particles in each atom~\citep{brown2009chemistry, halliday2013fundamentals}.
In addition, applying atom modeling in a neural network also amounts to regularizing the representation space to preserve each data sample's uniqueness.
Finally, atom modeling promotes expressiveness using a loss function with no latent supervision, enabling it to be flexibly applied to different applications.

We demonstrate the utility of atom modeling objective functions by training or finetuning convolution neural networks, generative adversarial networks, and transformers on Gaussian mixtures, natural texts (CoLA, Poem), and natural images (MNIST, CIFAR10, CelebA-HQ, Oxford-IIIT Pets, Oxford-Flowers102, ImageNet-1K).
Our experiments demonstrate that atom modeling outperforms baselines and provides an interpretation of how each sub-unit affects the learning, and shows how atom modeling alters the inter-sample relationship.

    \section{Related Work}

Atom modeling can be interpreted as a way of learning representation by spacing data samples.
The idea of keeping a distance among data samples has previously been used in manifold learning~\citep{tenenbaum2000global,saul2003think,cayton2005algorithms,lin2008riemannian}, graph representations~\citep{perozzi2014deepwalk,grover2016node2vec,hamilton2017representation}, kernel tricks~\citep{muller2001introduction,keerthi2003asymptotic,hofmann2008kernel}, and contrastive learning~\citep{weinberger2009distance,gutmann2010noise,sohn2016improved,oord2018representation,chen2020simple} where representations are trained to fit a predefined inter-sample relationship.
For instance, contrastive learning, the most related one to our method, requires preset negative and positive pairs in order to push away opposing pairs while bringing together the positive pairs.
These methods necessitate prior knowledge of inter-sample relationships.

Since atom modeling can be seen as a technique to discretize representations within a {\it continuous space}, it is natural to consider its {\it discrete space} counterpart: vector quantized variational autoencoder~\citep{van2017neural,razavi2019generating,esser2021taming}, which maps encoder output to an additional codebook with preset number of codes as the way of discretization.
This shows success in reconstruction but is not easy to generalize to other models.
In comparison, atom modeling that promotes separation of the original continuous embedding space frees the restriction on the preset number of codes and the autoencoder architecture.

While atom modeling leverages fine-grained component importance to determine the {\it balanced} data sample distances, it also provides model-agnostic partial interpretation that is orthogonal to extensions of model-dependent self-interpretable designs~\cite{alvarez2018towards} and post-hoc explanation methods~\cite{ribeiro2016should}.
However, this work does not focus on explanation but demonstrates an outgrowth of atom modeling.
    
    \section{Method}
Our goal is to define a flexible method that makes a model more expressive for data samples with multiple sub-units, such as images and texts, and does not need latent space supervision.
We say a model is expressive if it can accommodate various distinct outputs for different inputs.

We define a model in a general form:
\begin{equation}
    y = f_\theta(\mathbf{z}), \mathbf{z}\sim \mathcal{D}\,,
\end{equation}
where $\mathcal{D}$ is the data distribution or a random noise distribution.
We can easily fit models for practical applications, such as generation or classification, into this form:
$y$ is often a real vector $\mathbf{y}$ for generation and a probability distribution $P(Y)$ for classification.
If $f_\theta$ is expressive, different $\mathbf{z}$ is more likely to give different $\mathbf{y}$ or $P(Y)$. 
This distinction is desirable for promoting diversity in generation models~\cite{razavi2019generating} and encouraging entropy in classification models~\cite{dubey2018maximum}.

To achieve the goal of distinct outputs, we first write a model in its composite form: 
\begin{equation}
    f_\theta(\cdot) = o(\ell(\cdot)), 
\end{equation}
where $\ell(\cdot)\in \mathbb{R}^{N\times h}$ gives the latent representation of an input, and $o(\cdot)$ outputs the result given the latent representation.
$N$ is the number of sub-components, and $h$ is the dimension of the latent space.
An intuitive way to increase the probability that $f_\theta(\mathbf{z}^A)$ differs from $f_\theta(\mathbf{z}^B)$ is to let $\ell(\mathbf{z}^A)$ distance from $\ell(\mathbf{z}^B)$. 
Here, we show the properties of the output function, $o(\cdot)$, that lead to this concurrent increase. 

\begin{lemma}\label{lemma:output-space-bounds}
    A G-Lipschitz function $o(\cdot)$ and a K-Lipschitz inverse function of $o(\cdot)$ returns the output space distance such that:
    \begin{equation}
        \label{eq:output-space-bounds}
        K \|\mathbf{\mathbf{v}}-\mathbf{u}\| \leq \|o(\mathbf{v})-o(\mathbf{u})\| \leq G \|\mathbf{v}-\mathbf{u}\|\,,
    \end{equation}
    where $\mathbf{v}$ and $\mathbf{u}$ are any vector in the latent space.
\end{lemma}
Equation~\ref{eq:output-space-bounds} indicates that if the latent distance increases, the bounds of output distance also increase. All proofs in this paper are in appendix~\ref{apx:proofs}.

The next challenge is that, in a general case without latent space supervision, how we should set apart the latent variables produced by $\ell(\cdot)$.
We propose to \textit{dynamically} distance latent representations by separating the {\it currently close} variables and neglecting the {\it already distant} variables. 
Whether the variables are close or distant depends on their intra-sample structures. 
This leads us to first map a latent variable to a new embedding space by a learnable mapping function $\mathcal{A}(\cdot)$ such that:
\begin{equation}
    \left\{ (q_i,\mathbf{p}_i) \right\}_{i=1}^N  = \mathcal{A}\left( \ell(\mathbf{z}) \right)\,,
\end{equation}
where $q_i \in \mathbb{R}$ is the \textit{importance score} of the $i$-th row (token) in $\ell(\mathbf{z})$, and $\mathbf{p}_i \in \mathbb{R}^{h'}$ is the position of the same token in the new space.

\begin{figure}[t]
    \centering
    \includegraphics[width=.9\linewidth]{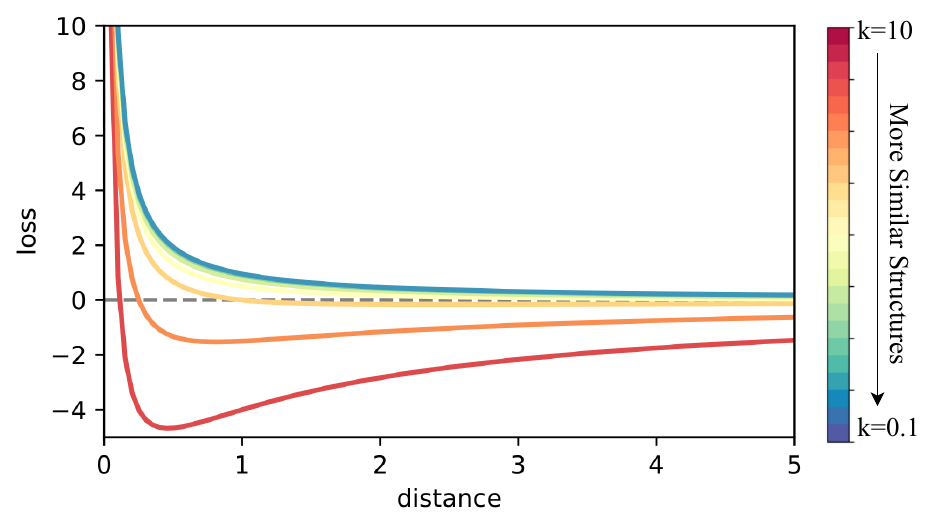}
    \caption{$\mathcal{L}_\mathcal{A}$ with varied atomic structure similarity $k$. The distance having the minimum loss depends on the intra-sample structures.
    As the structures are more similar (decayed $k$), the minimum loss distance becomes larger.
    Simultaneously, the distance cannot be zero.
    }
    \label{fig:atom-loss}
\end{figure}

Then, we propose a dynamic distancing loss function:
\begin{equation}
    \label{eq:loss}
    \mathcal{L}_\mathcal{A} = E_{\mathbf{z}^A,\mathbf{z}^B\sim \mathcal{D}} \sum_{\substack{i \in A,j \in B}}
    \frac{q_i^A q_j^B}{d\left( q_i^A, q_j^B, \mathbf{p}_i^A, \mathbf{p}_j^B \right)}, 
\end{equation}
where $d\left( q_i^A, q_j^B, \mathbf{p}_i^A, \mathbf{p}_j^B \right) \in \mathbb{R}$ is a {\it distance} between $i$-th and $j$-th tokens in $\mathbf{z}^A$ and $\mathbf{z}^B$ and is derived from their intra-sample structures. We also use $A$ and $B$ as sets $\{1,\cdots,N_A\}$ and $\{1,\cdots,N_B\}$.

By minimizing $\mathcal{L}_\mathcal{A}$ in Equation \ref{eq:loss}, the optimal distance between $\ell(\mathbf{z}^A)$ and $\ell(\mathbf{z}^B)$ cannot be $0$. 
That is, our proposed atomic loss forces $\ell(\mathbf{z}^A)$ and $\ell(\mathbf{z}^B)$ to be apart. 
In addition, the optimal distances are not identical for different data pairs, and these optimal values depend on each data's intra-sample structure. 
Figure \ref{fig:atom-loss} shows examples of the atomic loss function and optimal distances.

\subsection{Token Importance}
In Equation~\ref{eq:loss}, $q_i^A\in \mathbb{R}$ is a learnable importance score of a token in a data sample $A$.
Given a latent representation of $A$, $\ell(\mathbf{z}^A) = [\mathbf{e}_1^A \mathbf{e}_2^A ... \mathbf{e}_{N_A}^A] \in \mathbb{R}^{N_A\times h}$, we define the token importance as:
\begin{equation}
    q_i^A = 2\sigma(Q(\mathbf{e}_i^A))-1 \in [-1,1]\,,\label{eq:token-importance}
\end{equation}
where $\sigma(\cdot)$ is the sigmoid function, and $Q(\cdot):\mathbb{R}^h\mapsto \mathbb{R}$ maps the original h-dimension latent variable to an {\it unnormalized} importance score.
We rescale the score to $[-1,1]$ as it is a simple way to have three types of multiplication $q_iq_j$ needed in Equation~\ref{eq:loss}: polarity (negative), likeness (positive), and no effect (zero).
Since only when $q_i$ is not zero, $q_iq_j$ attributes to $\mathcal{L}_\mathcal{A}$, one role of token importance is as asking {\it if the i-th token in a data sample makes it distinguishable from other data samples}.
As shown in Table~\ref{tab:q-meaning}, token importance +1 or -1 helps distinguish data while 0 does not.

\begin{table}[t]\small
    \centering
    \begin{tabular}{cc|c|c}
        & & \multicolumn{2}{c}{\it Attribute to Meaning?}\\
        & & Yes & No\\\hline
        \multirow{2}{*}{\it Distinguish data?} & Yes & +1 & -1\\\cline{2-4}
        & No & 0& Not considered\\
    \end{tabular}
    \caption{Interpretation of token importance $q_i$.}
    \label{tab:q-meaning}
\end{table}

\subsection{Atomic Distance}

We define the {\it atomic distance} between data samples A and B by:
\begin{equation}\label{eq:atomic-distance}
    \bar{d}_{AB} = \|\boldsymbol{\mu}_A - \boldsymbol{\mu}_B\|_p\,,
\end{equation}
and the distances among their $i$-th and $j$-th tokens $d(q_i^A,q_j^B, \mathbf{p}_i^A, \mathbf{p}_j^B) \in \mathbb{R}$ in Equation~\ref{eq:loss} or $d_{ij}$ for brevity by:
\begin{equation}\label{eq:token distance}
    d_{ij} = \bar{d}_{AB} + \frac{r_A+r_B}{2} step(-q_i^Aq_j^B), \forall i\in A, j\in B\,.
\end{equation}
Here $\boldsymbol{\mu}_A$ and $\boldsymbol{\mu}_B$ are respectively an average of the most crucial tokens of A and B, $r_A$ and $r_B$ are the token {\it deviation} within a data sample, and $step(\cdot)$ is the step function.
We formally define them as:
\begin{align}
    \boldsymbol{\mu}_A & = \frac{1}{N_A} \sum_{i\in A} m_i^A \mathbf{p}_i^A\,,\\
    r_A & = \frac{1}{N_A} \sum_{i\in A} (1-m_i^A)\|\mathbf{p}_i^A - \boldsymbol{\mu}_A\|_p\,,\label{eq:radius}\\
    m_i^A & =1-\max(-q_i^A,0)\in [0,1]\,,\\
    \mathbf{p}_i^A & = \mathcal{P}(\mathbf{e}_i^A) \in \mathbb{R}^{h'}\,,\label{eq:atomic-position}
\end{align}
where $\mathcal{P}(\cdot):\mathbb{R}^h \mapsto \mathbb{R}^{h'}$ maps the original latent variable to the position of the $i$-th token in a new space, and $m_i^A\in \mathbb{R}$ is the {\it mass} of $i$-th token, indicating how much the token attributes to the meaning of A.
As shown in Table~\ref{tab:q-meaning}, tokens with $q_i$ being +1 or 0 play a key role in the data meaning and have $m_i$=1, while tokens with importance -1 are less likely to attribute to data meaning.

Since Equation~\ref{eq:loss} optimizes the atomic distance $\bar{d}_{AB}$ that is not a conventional distance metric, we show its relationship to Euclidean distance.
\begin{theorem}\label{theorem:distance-bound}
    Consider equal token importance distribution, Equation~\ref{eq:atomic-distance} returns the atomic distance such that:
    \begin{equation}
        \bar{d}_{AB} \leq C \|\mathbf{\tilde{v}}^A - \mathbf{\tilde{v}}^B\|_2\,,
    \end{equation}
    where $\mathbf{\tilde{v}}^A$ is a permutation of $\ell(\mathbf{z}^A)$ from data sample A.
\end{theorem}
Theorem~\ref{theorem:distance-bound} implies that rising $\bar{d}_{AB}$ encourages separation of $\ell(\mathbf{z}^A)$ and $\ell(\mathbf{z}^B)$ in the Euclidean space.
Therefore, according to Lemma~\ref{lemma:output-space-bounds}, rising $\bar{d}_{AB}$ can increase the bounds of $\left\lVert o(\ell(\mathbf{z}^A)) - o(\ell(\mathbf{z}^B)) \right\rVert_2$.

Next, we show that by optimizing Equation~\ref{eq:loss}, the distance between two data samples depends on how similar their atomic structures are.
In other words, the inter-sample relationship depends on the intra-sample structures.
This dependence is crucial to achieve dynamic distance among data samples.

\begin{theorem}\label{theorem-distance}
Let $c = {\sum}_{q_iq_j>0} q_iq_j$ and $c^* = {\sum}_{q_iq_j<0} q_iq_j$ for all $i\in A$ and $j\in B$, 
given data samples $A$, $B$.
Without loss of generality, $c^* = k c$ and $k\in(1,\infty)$ gives the optimal atomic distance in Equation~\ref{eq:loss} as:
\begin{equation}
   \bar{d}_{AB}^* = \bigl(\frac{r_A+r_B}{2}\bigr) \frac{\sqrt{k}+1}{k-1}
\end{equation}

If $k\rightarrow 1$, $\bar{d}_{AB}^* \rightarrow \infty$ and $k\rightarrow \infty$, $\bar{d}_{AB}^* \rightarrow 0$
\end{theorem}
Theorem \ref{theorem-distance} shows that optimizing Equation~\ref{eq:loss} forces data samples with similar intra-sample structures to separate more than the ones with dissimilar structures. 
Hence, our method results in \textit{dynamic distancing}.

\subsection{Training}
We further prevent the model from learning every tokens equally important using a soft constraint on the token importance distribution, which is essential to form reasonable intra- and inter-sample relationship. 
We regularize the number of tokens with different importance scores to be similar.
The complete loss function of Equation~\ref{eq:loss} becomes:
\begin{equation}\label{eq:all-loss}\small
    \mathcal{L}_\mathcal{A} = \underset{A,B\sim \mathcal{D}}{E} \sum_{i\in A,j\in B} \frac{q_i^Aq_j^B}{d_{ij}} + \left(\sum_{i\in A} q_i \right)^2 + \left(\sum_{i\in A} q_i^2 - \frac{2}{3} N_A\right)^2
\end{equation}

Algorithm~\ref{alg:atom-modeling} lists the complete training process.

\begin{algorithm}[tb]
   \caption{Atom Modeling}
   \label{alg:atom-modeling}
\begin{algorithmic}
   \STATE {\bfseries Input:} data $\mathcal{D}$, random distribution $\mathcal{R}$, model $f_\theta := o(h(\cdot))$, training criterion $\mathcal{L}_{ori}$, batch size $M$
   \FOR{$t=1$ {\bfseries to} Training Ends}
   \STATE $(z,y)\sim \mathcal{D}$
   \STATE Form batch $\mathcal{B} = \{(\mathbf{z}^b,y^b)\}_{b=1}^M$
   \STATE $\mathbf{e}^b = h(\mathbf{z}^b)$ and $\hat{y}^b = o(\mathbf{e}^b)$
   \STATE Get $\mathcal{L}_{ori}(y,\hat{y})$ for all $(z,y)\sim \mathcal{B}$
   \STATE Map $\mathbf{e}^A,\mathbf{e}^B$ to $q_i^A$, $q_j^B$ $d_{ij}$ as Eq~\ref{eq:token-importance}-\ref{eq:atomic-position} $\forall z^A,z^B\sim \mathcal{B}$
   \STATE Get $\mathcal{L}_{\mathcal{A}}(q_i^A,q_j^B,d_{ij})$ as Eq~\ref{eq:all-loss}
   \STATE Update $\theta$ by minimizing $\mathcal{L}_{ori} + \mathcal{L}_\mathcal{A}$
   \ENDFOR
\end{algorithmic}
\end{algorithm}

\begin{figure*}[t]
    \centering
    \includegraphics[width=.85\linewidth]{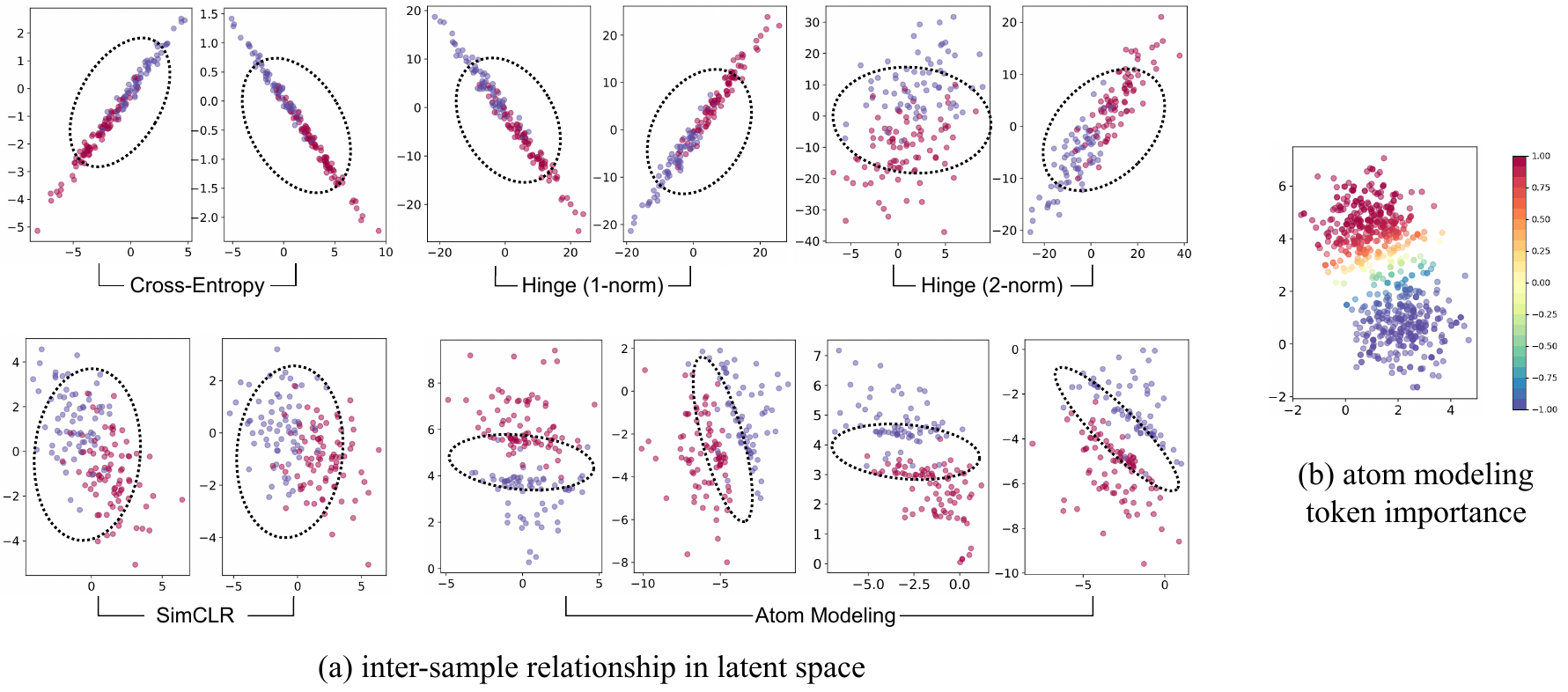}
    \caption{(a) Visualization of the latent space of synthetic data by only cross-entropy training loss or integrated with hinge losses (L1 and L2), SimCLR, or Atom Modeling. Blue and red indicates the two ground-truth classes. The dashed circles annotate the overlaps of the learned representaions from different classes, which is correlated with the easiness to classify the samples. Atom modeling separates representations with a gap using no latent supervision. (b) Visualization of token importance.}
    \label{fig:inter-structure-synth-exp}
\end{figure*}

\subsection{Relation to Atomic Physics.}
Our proposed method has high correspondence with atomic physics~\citep{halliday2013fundamentals,brown2009chemistry}, the scientific study of the structure of an atom and its interaction with others.
Therefore, we name this method {\it atom modeling}.

Among atoms in nature, there are {\it inter-atomic} forces, similar to our proposed loss function in Equation~\ref{eq:loss}, that bind atoms and avoid them collapsing by maintaining a balanced distance.
The distances among a group of atoms depend on their {\it atomic structures}, similar to our theoretical result in Theorem~\ref{theorem-distance}.

Within an atom, its structure consists of three types of particles: neutrons, protons, and electrons, where a neutron has no charge, a proton has one positive charge with similar weight as a neutron, and an electron has a negative charge and weight significantly less than a proton~\citep{mohr2008codata}.
The charges correspond to our token importance.

Simultaneously, \citet{bohr1913constitution} introduced one way to describe an atomic structure, where the protons and neutrons form a {\it nucleus}, similar to $\mathbf{\mu}$, that occupies a small volume of the atom while the electrons orbit around the nucleus with a {\it radius}, similar to our $r$.

The soft constraint in Equation~\ref{eq:all-loss} is also similar to that the protons, electrons, and neutrons often have similar numbers within one atom.

    \section{Experiments}\label{sec:exp}
We test atom modeling's effects and flexibility by training linear classifiers on synthetic data, GANs on unconditional image generation, ResNets on image classification, and transformers on text classification.

In the experiments, while $Q(\cdot)$ and $\mathcal{P}(\cdot)$ can be any mapping functions, we use simple extraction functions with a selected hidden layer such that $Q(\cdot)$ extracts one dimension from the original $\ell(\cdot)\in\mathbb{R}^h$ and $\mathcal{P}(\cdot)$ extracts the rest $h-1$ dimensions.
More discussion is in appendix~\ref{apx:discussion}.

\subsection{Linear Classifier on Synthetic Data}
To demonstrate the shifts of inter-sample relationships after atom modeling, we first conduct experiments on synthetic data.
We mimic data of multiple sub-components by generating input features $X$ and the corresponding labels $y$ as follows:
\begin{equation}
    \left\{
    \begin{array}{l}
        P(A) = \mathcal{N}(\mu_a, \sigma_a)\\
        P(B) = \mathcal{N}(\mu_b, \sigma_b)\\
        X = \{x_i | x_i \in A\cup B\}_{i=1}^N\\
        y = \mathbbm{1}\left( P(x_i\in A | x_i \in X) > P(x_i\in B | x_i \in X)\right)\\
    \end{array}
    \right.
\end{equation}
where $A$ and $B$ are two events of normal distributions with ($\mu_a$, $\sigma_a)$ and ($\mu_b$, $\sigma_b$) being the mean and standard deviation respectively.
$X$ is the input composed of $N=5$ sub-units $x_i$ sampled from $A\cup B$. The goal is to find a function $f:X\rightarrow y$.

\begin{figure}[t]
    \centering
    \includegraphics[width=.9\linewidth]{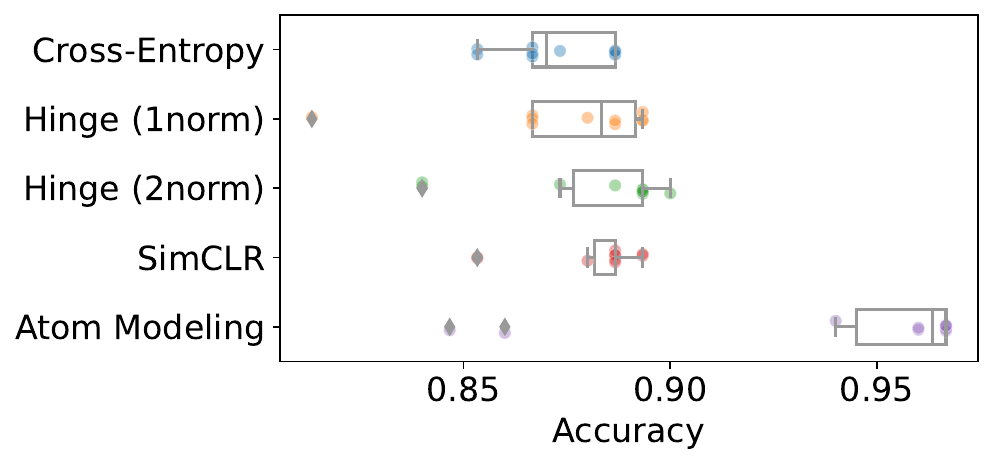}
    \caption{Comparison among cross-entropy, p-norm distance, SimCLR, and atom modeling.}
    \label{fig:synth-exp-box}
\end{figure}

In this experiment, we use a neural network with two fully-connected linear layers and apply atom modeling to the hidden state of the first layer.
For comparison, we employ Hinge loss with p-norm distances~\cite{bromley1993signature,chopra2005learning,lecun2006tutorial} and SimCLR~\citep{chen2020simple} that uses cosine similarity as the metric on the same hidden state, as our baselines.
Figure~\ref{fig:synth-exp-box} shows the classification accuracy across ten random runs.
Atom modeling enhances the classifier to achieve an average of 96\% accuracy and is superior to baselines.

\begin{figure*}[t]
    \centering
    \includegraphics[width=.9\linewidth]{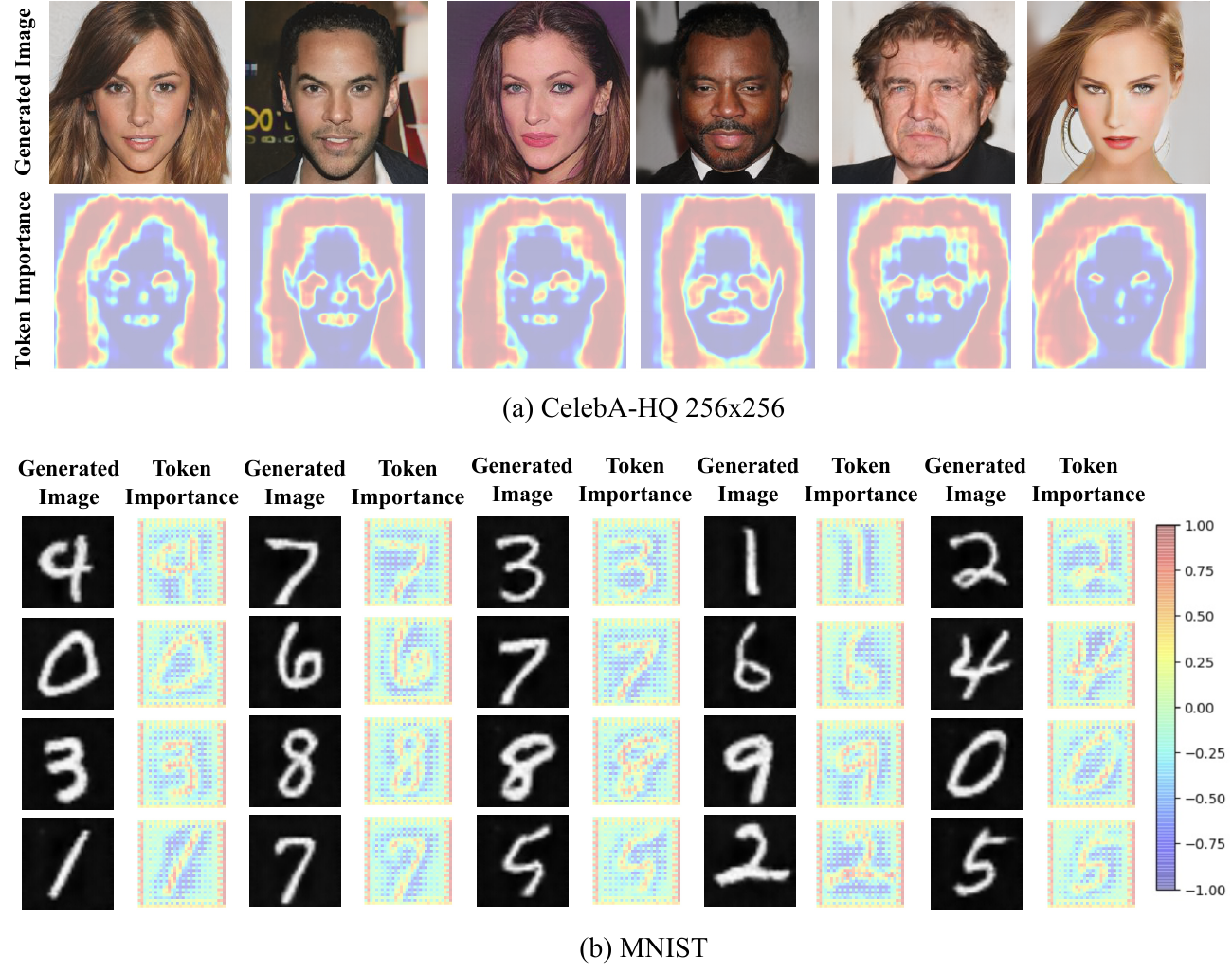}
    \caption{Examples of generated images and the learned token importance by atom modeling on unconditional image generation. The distributions show that importance score close to one indicates it is a crucial part of the image to distinguish from others.}
    \label{fig:img-gen-charges}
\end{figure*}
\begin{figure*}
    \centering
    \includegraphics[width=.8\linewidth]{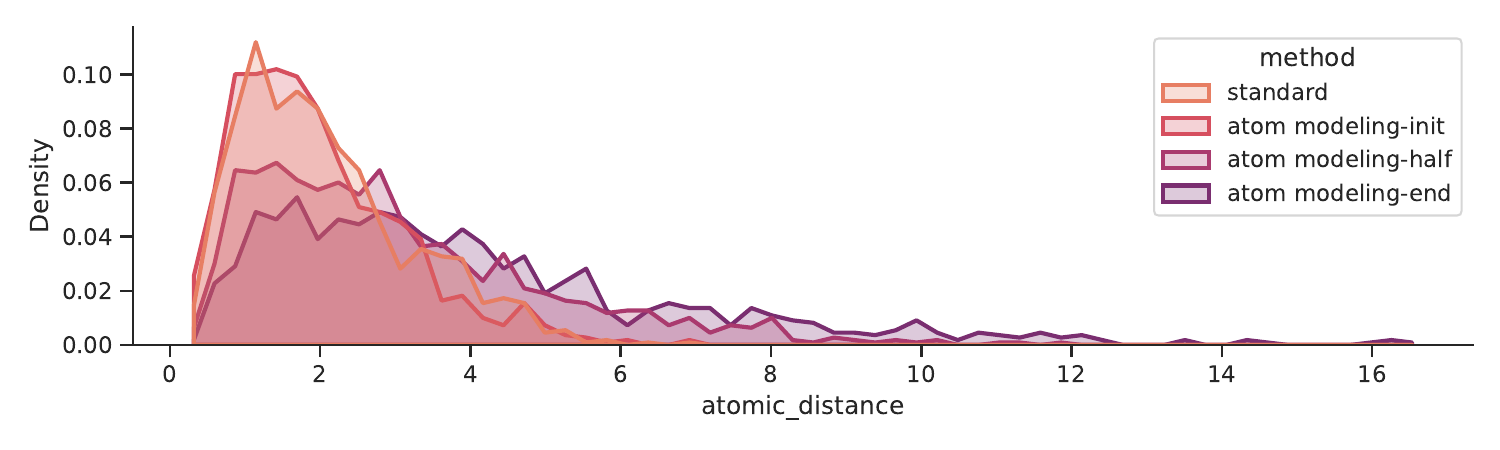}
    \caption{Alteration of atomic distance distributions by atom modeling (initial stage, half, end of training) and comparison with standard training criterion. Atom modeling gradually disseminates the atomic distance distribution.}
    \label{fig:mnist-atom-dist}
\end{figure*}

We visualize how atom modeling alters inter-sample relationships and learns token importance.
Figure~\ref{fig:inter-structure-synth-exp}(a) demonstrates that atom modeling spreads out the representation distribution, especially the high-density region.
This further creates a gap between the blue and red classes in the latent space, thus enhancing the classifier expressivity.
The corresponding token importance is plotted in Figure~\ref{fig:inter-structure-synth-exp}(b).
The model takes sub-units near the $\mu_a$ ($q=+1$) and $\mu_b$ ($q=-1$) as the most crucial ones to distinguish data samples and takes sub-units near $\mu_a$ ($q=+1$) and $(\mu_a+\mu_b)/2$ ($q=0$) as the keys to data meaning.

\subsection{GANs on Unconditional Image Generation}

We investigate atom modeling on generative models with unconditional image synthesis tasks: MNIST~\citep{lecun1998gradient}, CIFAR10~\citep{krizhevsky2009learning}, and CelebA-HQ256x256~\citep{karras2018progressive}.
For MNIST and CIFAR10, we performed experiments with DCGAN~\cite{radford2015unsupervised}.
For CelebA-HQ256x256, we performed experiments with the SOTA model, StyleSwin~\cite{zhang2022styleswin}, and followed their implementation.

\begin{table}[t]\small
    \centering
    \caption{FIDs of image synthetic on MNIST, CIFAR10, and CelebA-HQ256x256.}
    \label{tab:results-imggen}
    \begin{tabular}{llc}\toprule[1pt]
         Method & Dataset & FID \\\midrule[0.5pt]
         DCGAN~\citep{radford2015unsupervised} & MNIST & 88.4 \\
         VQ~\citep{van2017neural} & MNIST & 303.9 \\
         SSCL~\citep{chen2020simple} & MNIST & 69.4 \\
         Atom Modeling & MNIST & \bf 49.0 \\\midrule[0.5pt]
         DCGAN~\citep{radford2015unsupervised} & CIFAR10 & 110.4 \\
         VQ~\citep{van2017neural} & CIFAR10 & - \\
         SSCL~\citep{chen2020simple} & CIFAR10 & 130.0 \\
         Atom Modeling & CIFAR10 & \bf 97.4 \\
         \midrule[0.5pt]
         VQGAN~\citep{esser2021taming} & CelebA-HQ & 10.2\\
         StyleSwin~\citep{zhang2022styleswin} & CelebA-HQ & 5.26\\
         Atom Modeling & CelebA-HQ & \bf 5.18\\
         \bottomrule[1pt]
    \end{tabular}
\end{table}

For comparison, we employed self-supervised contrastive learning~\cite{chen2020simple} and vector quantization~\cite{van2017neural} as additional loss to regularize a given representation layer and compared their ability with atom modeling for end-to-end training.
In this experiment, for DCGAN, we use the output from the last hidden layer of the generator as the representation to be discretized.
For StyleSwin, we use the output from the last hidden layer with half resolution.
All the generated results are evaluated by Fr\'echet Inception Distance (FID)~\citep{heusel2017gans, obukhov2020torchfidelity} with 2k images for MNIST, 10k for CIFAR10, and 30k for CelebA-HQ256x256.
Lower FID indicates higher generation quality.

Empirically, Table~\ref{tab:results-imggen} demonstrates the effectiveness of atom modeling.
The proposed atom modeling outperforms VQ- or SSCL-enhanced DCGAN on MNIST (49.0 vs 69.4) and CIFAR10 (97.4 vs 110.4). Additionally, our approach improves StyleSwin on CelebA-HQ256x256 in our reproduced results (5.18 vs 5.26).
Note that this promising improvement is gained under the original setup of DCGAN and StyleSwin without extra tuning.
This shows that atom modeling can improve generative model expressivity in flexible settings.

Figure~\ref{fig:img-gen-charges} shows examples of our generated images and the associated token importance after applying atom modeling.
Areas mapped to hair, eyes, nose, and philtrum in CelebA-HQ256x256, as well as the cores of digits in MNIST, have positive importance scores.
They play a crucial role in both distinguishing from other images and semantic meaning.
Areas map to skin and background have negative importance scores.
They are pivotal for differing from others but not the meaning.
Other regions are less likely to identify an image but attribute to the semantic meaning.
They are, therefore, assigned importance scores of zero.

We plot the atomic distance distribution over training time in Figure~\ref{fig:mnist-atom-dist}.
At the initial training stage using atom modeling, the distances among data samples are similar to standard training results.
The distances in the latent space concentrate to a small value.
During training, atom modeling modifies the latent space and gradually disseminates the distance distribution.
This matches our expectation of what atom modeling has done during model learning.

\subsection{ResNet on Image Classification}
\begin{figure}
    \centering
    \includegraphics[width=.8\linewidth]{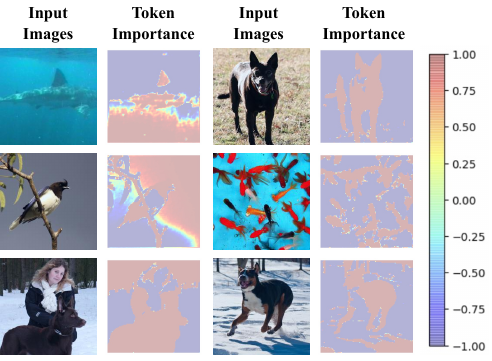}
    \caption{Visualization of the learned charges by atom modeling on unconditional image synthesis (CelebA-HQ 256x256, MNIST) and image classification (ImageNet-1K). The distributions show that protons (charge close to +1) are often the crucial parts in an image to be distinguished from others.}
    \label{fig:img-cls-charges}
\end{figure}

\begin{figure*}[t]
    \centering
    \includegraphics[width=.9\linewidth]{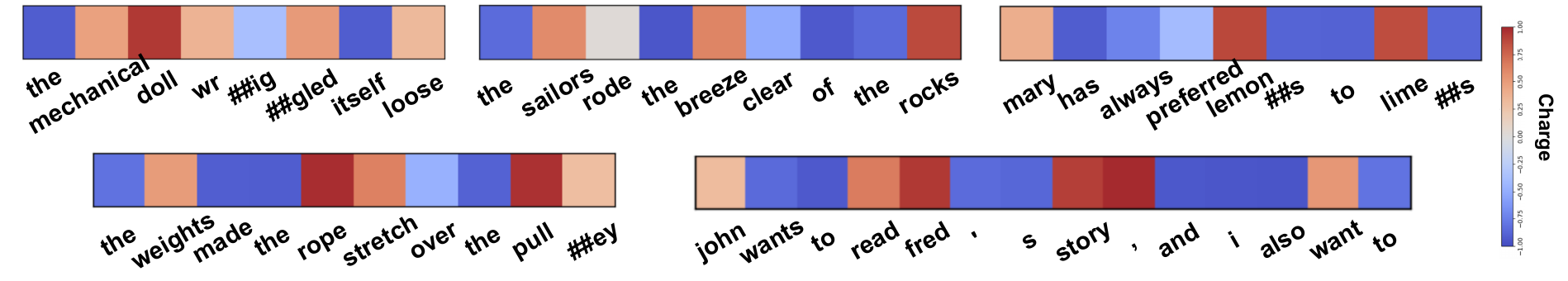}
    \caption{Visualization of the learned charges by atomic modeling on text classification (COLA). The charge distributions again show that protons are often mapped to the keywords in a sentence.}
    \label{fig:language-charges}
\end{figure*}

We also validate atom modeling on fine-grained image classification Oxford-IIIT Pets~\cite{parkhi2012cats} and Oxford-Flowers102~\cite{nilsback2008automated} as well as ImageNet-1K~\cite{deng2009imagenet} to justify its flexibility.

In a fine-grained classification problem, intra-class diversity is higher than inter-class diversity~\cite{wei2019deep}, so we can use higher expressivity and, thus, atom modeling.
\begin{table}[t]\small
    \centering
    \caption{Results of fine-grained classification on Oxford-IIIT Pets, Oxford-Flowers102, CoLA, and Poem datasets.}
    \label{tab:results-fine-grained}
    \begin{tabular}{@{} lcc @{}}\toprule[1pt]
        & Pets & Flowers\\
        Method & Acc & Acc\\\midrule[0.5pt]
        
        Cross-Entropy
        & 21.0 & 56.7\\
        Hinge (1-norm)~\citep{lecun2006tutorial}
        & 20.0 & 54.7\\
        Hinge (2-norm)~\citep{chopra2005learning}
        & 22.4 & 58.0\\
        Rank-H~\citep{wang2015unsupervised}
        & 22.4 & 58.2\\
        SimCLR~\citep{chen2020simple}
        & 20.7 & 58.1\\
        Atom Modeling
        & \bf 22.5 & \bf 59.1\\
        
        \bottomrule[1pt]
    \end{tabular}
\end{table}

\begin{table}[t]\small
    \caption{Results of ImageNet-1K with \citep{he2016deep,khosla2020supervised} data augmentation approaches.}
    \label{tab:results-cls-imagenet}
    \begin{tabular}{@{} lcc @{}}\toprule[1pt]
        Method & Top-1 & Top-5\\\midrule[0.5pt]
        Cross-Entropy w/ \citep{he2016deep} & 74.97 & 92.17 \\
        Atom Modeling w/ \citep{he2016deep} & \bf 75.10 & \bf 92.25 \\\midrule[0.5pt]
        Cross-Entropy w/ \citep{khosla2020supervised} & 75.02 & 92.20 \\
        Atom Modeling w/ \citep{khosla2020supervised} & \bf 75.19 & \bf 92.35 \\
        \bottomrule[1pt]
    \end{tabular}
\end{table}

For comparison, we train ResNet18~\cite{he2016deep} with cross-entropy loss, while employing hinge loss with p-norm distances, Rank-H~\cite{wang2015unsupervised}, SimCLR~\cite{chen2020simple}.
In Table~\ref{tab:results-fine-grained}, we present the mean top-1 accuracy (Acc).
The empirical results show that atom modeling consistently improves cross-entropy and is the best among distancing-representation-like approaches.

We further examine the ability of atom modeling applied to a larger-scale general classification problem on ImageNet-1K.
We follow prior work implementations to use ResNet50~\citep{he2016deep} as the backbone and run 90 epochs with two data augmentation methods used in~\citep{he2016deep, khosla2020supervised}. The first includes only the crop and horizontal flip, and the second adds color jitters and grayscale.
Table~\ref{tab:results-cls-imagenet} shows that atom modeling improves cross-entropy under different data augmentations, which has been found to impact the results of image classification~\citep{cubuk2019autoaugment,cubuk2020randaugment,khosla2020supervised}.
Note that this performance gain has only been made by introducing atom modeling to one representation layer and using the exact same setup of conventional training of ResNet50 on ImageNet-1K; with a more elaborate setting, the performance could be improved.
More importantly, the outcomes show that atom modeling can be flexibly applied to diverse data, models, and scales.

\begin{table}[t]\small
    \centering
    \caption{Results of fine-grained classification on Oxford-IIIT Pets, Oxford-Flowers102, CoLA, and Poem datasets.}
    \label{tab:results-fine-grained}
    \begin{tabular}{@{} lcc @{}}\toprule[1pt]
        
        & CoLA & Poem\\
        Method & MCC & F1\\\midrule[0.5pt]
        Cross-Entropy
        & 60.0 & 60.3\\
        Hinge (1-norm)~\citep{lecun2006tutorial}
        & 59.8 & 62.0\\
        Hinge (2-norm)~\citep{chopra2005learning}
        & 60.1 & 61.8\\
        SimCSE~\citep{gao2021simcse}
        & 60.8 & 59.3\\
        MixCSE~\citep{zhang2022unsupervised}
        & 60.4 & 60.8\\
        Atom Modeling &
        \bf 61.3 & \bf 62.7\\
        
        \bottomrule[1pt]
    \end{tabular}
\end{table}

\subsection{Transformer on Text Classification}

We further experimented on fine-grained text classification: CoLA~\citep{wang2018glue, warstadt-etal-2019-neural} and Poem~\citep{sheng2020investigating}.
For comparison, we finetuned BERT~\cite{devlin2019bert} for language with cross-entropy loss, while employing hinge loss with p-norm distances, SimCSE~\cite{gao2021simcse}, and MixCSE~\cite{zhang2022unsupervised}.
In Table~\ref{tab:results-fine-grained}, we present the Matthews's correlation coefficients (MCC) and F1 scores as used in prior work for each task.
The empirical results show that our method consistently improves cross-entropy and is superior to the baselines.

The trained intra-sample relationship shows similarity to the vision domain.
In Figure~\ref{fig:language-charges}, we observe that tokens with special meanings have positive importance scores.
They contribute to both the uniqueness and semantics of the sentence.
The often-seen tokens, such as prepositions and articles, have negative importance scores.
They contribute to distinguishing some sentences but less the semantics.
Visualizability of the learned token importance exhibits the partial interpretability provided by atom modeling without post-hoc processing~\cite{ribeiro2016should}.

\subsection{Ablation Study}

\begin{figure}[t]
    \centering
    \includegraphics[width=.9\linewidth]{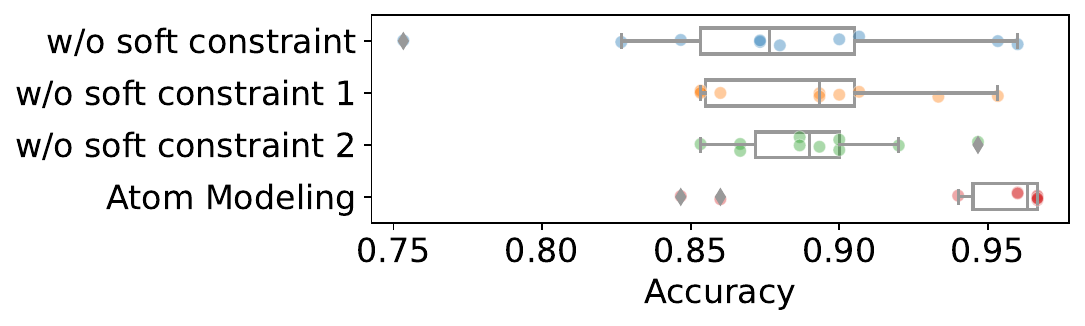}
    \caption{Ablation study of the soft constraint on linear classifiers of Gaussian mixtures with ten random runs.}
    \label{fig:ablation-synthetic-exp}
\end{figure}

\begin{table}[t]\small
    \centering
    \begin{tabular}{c|cc|cc}
         & \multicolumn{2}{|c}{MNIST} & \multicolumn{2}{|c}{CIFAR10}\\
         ablation & best & average & best & average\\\hline
         w/o constraint & -7.6 & -29.4 & -11.2 & -16.5\\
         w/o constraint 1 & -25.2 & -35.3 & -21.4 & -10.9\\
         w/o constraint 2 & -2.6 & -37.3 & -19.9 & -15.5\\
    \end{tabular}
    \caption{Ablation study of the soft constraint on GANs of MNIST and CIFAR10 image synthesis FID with five random runs.}
    \label{tab:ablation-image-synthesis}
\end{table}

We studied the impacts of the soft constraint $\left(\sum_{i\in A} q_i \right)^2 + \left(\sum_{i\in A} q_i^2 - \frac{2}{3} N_A\right)^2$ as shown in Figure~\ref{fig:ablation-synthetic-exp} and Table~\ref{tab:ablation-image-synthesis}.
We observe that only using the term $E_{A,B\sim \mathcal{D}} \sum_{i\in A,j\in B} \frac{q_iq_j}{d_{ij}}$, in most cases, can achieve good performance but suffers from high variance.
When having the soft constraint, the training performance is stabilized.

    \section{Conclusion}

We have presented atom modeling, a new algorithm to achieve model expressiveness by (1) learning token importance for each sub-unit in a data sample, and (2) dynamically distancing data samples based on their structural similarity with no supervision.
The learned token importance may be of independent interests as a partial model explanation.
Our method is also highly practical, demonstrating effectiveness across diverse deep learning problems.

\section*{Potential Broader Impact}
\label{apx:broader-impact}
This paper presents work whose goal is to advance the field of Machine Learning. There are many potential societal consequences of our work, none which we feel must be specifically highlighted here.

\bibliography{main}

\begin{thebibliography}{57}
\providecommand{\natexlab}[1]{#1}
\providecommand{\url}[1]{\texttt{#1}}
\expandafter\ifx\csname urlstyle\endcsname\relax
  \providecommand{\doi}[1]{doi: #1}\else
  \providecommand{\doi}{doi: \begingroup \urlstyle{rm}\Url}\fi

\bibitem[Alvarez-Melis \& Jaakkola(2018)Alvarez-Melis and Jaakkola]{alvarez2018towards}
Alvarez-Melis, D. and Jaakkola, T.~S.
\newblock Towards robust interpretability with self-explaining neural networks.
\newblock In \emph{NeurIPS}, 2018.

\bibitem[Bastani et~al.(2017)Bastani, Kim, and Bastani]{bastani2017interpreting}
Bastani, O., Kim, C., and Bastani, H.
\newblock Interpreting blackbox models via model extraction.
\newblock \emph{arXiv preprint arXiv:1705.08504}, 2017.

\bibitem[Bohr(1913)]{bohr1913constitution}
Bohr, N.
\newblock I. on the constitution of atoms and molecules.
\newblock \emph{The London, Edinburgh, and Dublin Philosophical Magazine and Journal of Science}, 26\penalty0 (151):\penalty0 1--25, 1913.

\bibitem[Bromley et~al.(1993)Bromley, Guyon, LeCun, S{\"a}ckinger, and Shah]{bromley1993signature}
Bromley, J., Guyon, I., LeCun, Y., S{\"a}ckinger, E., and Shah, R.
\newblock Signature verification using a" siamese" time delay neural network.
\newblock \emph{Advances in neural information processing systems}, 6, 1993.

\bibitem[Brown(2009)]{brown2009chemistry}
Brown, T.~L.
\newblock \emph{Chemistry: the central science}.
\newblock Pearson Education, 2009.

\bibitem[Cayton(2005)]{cayton2005algorithms}
Cayton, L.
\newblock Algorithms for manifold learning.
\newblock \emph{Univ. of California at San Diego Tech. Rep}, 12\penalty0 (1-17):\penalty0 1, 2005.

\bibitem[Chen et~al.(2020)Chen, Kornblith, Norouzi, and Hinton]{chen2020simple}
Chen, T., Kornblith, S., Norouzi, M., and Hinton, G.
\newblock A simple framework for contrastive learning of visual representations.
\newblock In \emph{International conference on machine learning}, pp.\  1597--1607. PMLR, 2020.

\bibitem[Chopra et~al.(2005)Chopra, Hadsell, and LeCun]{chopra2005learning}
Chopra, S., Hadsell, R., and LeCun, Y.
\newblock Learning a similarity metric discriminatively, with application to face verification.
\newblock In \emph{2005 IEEE Computer Society Conference on Computer Vision and Pattern Recognition (CVPR'05)}, volume~1, pp.\  539--546. IEEE, 2005.

\bibitem[Coulomb(1785)]{coulomb1785premier}
Coulomb, C.~A.
\newblock Premier m{\'e}moire sur l’{\'e}lectricit{\'e} et le magn{\'e}tisme.
\newblock \emph{Histoire de l’Academie royale des sciences}, 569, 1785.

\bibitem[Cubuk et~al.(2019)Cubuk, Zoph, Mane, Vasudevan, and Le]{cubuk2019autoaugment}
Cubuk, E.~D., Zoph, B., Mane, D., Vasudevan, V., and Le, Q.~V.
\newblock Autoaugment: Learning augmentation strategies from data.
\newblock In \emph{Proceedings of the IEEE/CVF conference on computer vision and pattern recognition}, pp.\  113--123, 2019.

\bibitem[Cubuk et~al.(2020)Cubuk, Zoph, Shlens, and Le]{cubuk2020randaugment}
Cubuk, E.~D., Zoph, B., Shlens, J., and Le, Q.~V.
\newblock Randaugment: Practical automated data augmentation with a reduced search space.
\newblock In \emph{Proceedings of the IEEE/CVF conference on computer vision and pattern recognition workshops}, pp.\  702--703, 2020.

\bibitem[Deng et~al.(2009)Deng, Dong, Socher, Li, Li, and Fei-Fei]{deng2009imagenet}
Deng, J., Dong, W., Socher, R., Li, L.-J., Li, K., and Fei-Fei, L.
\newblock Imagenet: A large-scale hierarchical image database.
\newblock In \emph{2009 IEEE conference on computer vision and pattern recognition}, pp.\  248--255. Ieee, 2009.

\bibitem[Devlin et~al.(2019)Devlin, Chang, Lee, and Toutanova]{devlin2019bert}
Devlin, J., Chang, M.-W., Lee, K., and Toutanova, K.
\newblock Bert: Pre-training of deep bidirectional transformers for language understanding.
\newblock In \emph{Proceedings of the 2019 Conference of the North American Chapter of the Association for Computational Linguistics: Human Language Technologies, Volume 1 (Long and Short Papers)}, pp.\  4171--4186, 2019.

\bibitem[Dosovitskiy et~al.(2014)Dosovitskiy, Springenberg, Riedmiller, and Brox]{dosovitskiy2014discriminative}
Dosovitskiy, A., Springenberg, J.~T., Riedmiller, M., and Brox, T.
\newblock Discriminative unsupervised feature learning with convolutional neural networks.
\newblock \emph{Advances in neural information processing systems}, 27, 2014.

\bibitem[Dubey et~al.(2018)Dubey, Gupta, Raskar, and Naik]{dubey2018maximum}
Dubey, A., Gupta, O., Raskar, R., and Naik, N.
\newblock Maximum-entropy fine grained classification.
\newblock \emph{Advances in neural information processing systems}, 31, 2018.

\bibitem[Esser et~al.(2021)Esser, Rombach, and Ommer]{esser2021taming}
Esser, P., Rombach, R., and Ommer, B.
\newblock Taming transformers for high-resolution image synthesis.
\newblock In \emph{Proceedings of the IEEE/CVF conference on computer vision and pattern recognition}, pp.\  12873--12883, 2021.

\bibitem[Gao et~al.(2021)Gao, Yao, and Chen]{gao2021simcse}
Gao, T., Yao, X., and Chen, D.
\newblock Simcse: Simple contrastive learning of sentence embeddings.
\newblock In \emph{Proceedings of the 2021 Conference on Empirical Methods in Natural Language Processing}, pp.\  6894--6910, 2021.

\bibitem[Grover \& Leskovec(2016)Grover and Leskovec]{grover2016node2vec}
Grover, A. and Leskovec, J.
\newblock node2vec: Scalable feature learning for networks.
\newblock In \emph{Proceedings of the 22nd ACM SIGKDD international conference on Knowledge discovery and data mining}, pp.\  855--864, 2016.

\bibitem[Gutmann \& Hyv{\"a}rinen(2010)Gutmann and Hyv{\"a}rinen]{gutmann2010noise}
Gutmann, M. and Hyv{\"a}rinen, A.
\newblock Noise-contrastive estimation: A new estimation principle for unnormalized statistical models.
\newblock In \emph{Proceedings of the thirteenth international conference on artificial intelligence and statistics}, pp.\  297--304. JMLR Workshop and Conference Proceedings, 2010.

\bibitem[Halliday et~al.(2013)Halliday, Resnick, and Walker]{halliday2013fundamentals}
Halliday, D., Resnick, R., and Walker, J.
\newblock \emph{Fundamentals of physics}.
\newblock John Wiley \& Sons, 2013.

\bibitem[Hamilton et~al.(2017)Hamilton, Ying, and Leskovec]{hamilton2017representation}
Hamilton, W.~L., Ying, R., and Leskovec, J.
\newblock Representation learning on graphs: Methods and applications.
\newblock \emph{arXiv preprint arXiv:1709.05584}, 2017.

\bibitem[He et~al.(2016)He, Zhang, Ren, and Sun]{he2016deep}
He, K., Zhang, X., Ren, S., and Sun, J.
\newblock Deep residual learning for image recognition.
\newblock In \emph{Proceedings of the IEEE conference on computer vision and pattern recognition}, pp.\  770--778, 2016.

\bibitem[Heusel et~al.(2017)Heusel, Ramsauer, Unterthiner, Nessler, and Hochreiter]{heusel2017gans}
Heusel, M., Ramsauer, H., Unterthiner, T., Nessler, B., and Hochreiter, S.
\newblock Gans trained by a two time-scale update rule converge to a local nash equilibrium.
\newblock \emph{Advances in neural information processing systems}, 30, 2017.

\bibitem[Hofmann et~al.(2008)Hofmann, Sch{\"o}lkopf, and Smola]{hofmann2008kernel}
Hofmann, T., Sch{\"o}lkopf, B., and Smola, A.~J.
\newblock Kernel methods in machine learning.
\newblock \emph{The annals of statistics}, 36\penalty0 (3):\penalty0 1171--1220, 2008.

\bibitem[Jain \& Wallace(2019)Jain and Wallace]{jain2019attention}
Jain, S. and Wallace, B.~C.
\newblock Attention is not explanation.
\newblock In \emph{Proceedings of the 2019 Conference of the North American Chapter of the Association for Computational Linguistics: Human Language Technologies, Volume 1 (Long and Short Papers)}, pp.\  3543--3556, 2019.

\bibitem[Karras et~al.(2018)Karras, Aila, Laine, and Lehtinen]{karras2018progressive}
Karras, T., Aila, T., Laine, S., and Lehtinen, J.
\newblock Progressive growing of {GAN}s for improved quality, stability, and variation.
\newblock In \emph{International Conference on Learning Representations}, 2018.
\newblock URL \url{https://openreview.net/forum?id=Hk99zCeAb}.

\bibitem[Keerthi \& Lin(2003)Keerthi and Lin]{keerthi2003asymptotic}
Keerthi, S.~S. and Lin, C.-J.
\newblock Asymptotic behaviors of support vector machines with gaussian kernel.
\newblock \emph{Neural computation}, 15\penalty0 (7):\penalty0 1667--1689, 2003.

\bibitem[Khosla et~al.(2020)Khosla, Teterwak, Wang, Sarna, Tian, Isola, Maschinot, Liu, and Krishnan]{khosla2020supervised}
Khosla, P., Teterwak, P., Wang, C., Sarna, A., Tian, Y., Isola, P., Maschinot, A., Liu, C., and Krishnan, D.
\newblock Supervised contrastive learning.
\newblock \emph{Advances in neural information processing systems}, 33:\penalty0 18661--18673, 2020.

\bibitem[Krizhevsky et~al.(2009)Krizhevsky, Hinton, et~al.]{krizhevsky2009learning}
Krizhevsky, A., Hinton, G., et~al.
\newblock Learning multiple layers of features from tiny images.
\newblock 2009.

\bibitem[LeCun et~al.(1998)LeCun, Bottou, Bengio, and Haffner]{lecun1998gradient}
LeCun, Y., Bottou, L., Bengio, Y., and Haffner, P.
\newblock Gradient-based learning applied to document recognition.
\newblock \emph{Proceedings of the IEEE}, 86\penalty0 (11):\penalty0 2278--2324, 1998.

\bibitem[LeCun et~al.(2006)LeCun, Chopra, Hadsell, Ranzato, and Huang]{lecun2006tutorial}
LeCun, Y., Chopra, S., Hadsell, R., Ranzato, M., and Huang, F.
\newblock A tutorial on energy-based learning.
\newblock \emph{Predicting structured data}, 1\penalty0 (0), 2006.

\bibitem[Lin \& Zha(2008)Lin and Zha]{lin2008riemannian}
Lin, T. and Zha, H.
\newblock Riemannian manifold learning.
\newblock \emph{IEEE transactions on pattern analysis and machine intelligence}, 30\penalty0 (5):\penalty0 796--809, 2008.

\bibitem[Mohr et~al.(2008)Mohr, Taylor, and Newell]{mohr2008codata}
Mohr, P.~J., Taylor, B.~N., and Newell, D.~B.
\newblock Codata recommended values of the fundamental physical constants: 2006.
\newblock \emph{Journal of Physical and Chemical Reference Data}, 80\penalty0 (3):\penalty0 633--1284, 2008.

\bibitem[Muller et~al.(2001)Muller, Mika, Ratsch, Tsuda, and Scholkopf]{muller2001introduction}
Muller, K.-R., Mika, S., Ratsch, G., Tsuda, K., and Scholkopf, B.
\newblock An introduction to kernel-based learning algorithms.
\newblock \emph{IEEE transactions on neural networks}, 12\penalty0 (2):\penalty0 181--201, 2001.

\bibitem[Nilsback \& Zisserman(2008)Nilsback and Zisserman]{nilsback2008automated}
Nilsback, M.-E. and Zisserman, A.
\newblock Automated flower classification over a large number of classes.
\newblock In \emph{2008 Sixth Indian Conference on Computer Vision, Graphics \& Image Processing}, pp.\  722--729. IEEE, 2008.

\bibitem[Obukhov et~al.(2020)Obukhov, Seitzer, Wu, Zhydenko, Kyl, and Lin]{obukhov2020torchfidelity}
Obukhov, A., Seitzer, M., Wu, P.-W., Zhydenko, S., Kyl, J., and Lin, E. Y.-J.
\newblock High-fidelity performance metrics for generative models in pytorch, 2020.
\newblock URL \url{https://github.com/toshas/torch-fidelity}.
\newblock Version: 0.3.0, DOI: 10.5281/zenodo.4957738.

\bibitem[Oord et~al.(2018)Oord, Li, and Vinyals]{oord2018representation}
Oord, A. v.~d., Li, Y., and Vinyals, O.
\newblock Representation learning with contrastive predictive coding.
\newblock \emph{arXiv preprint arXiv:1807.03748}, 2018.

\bibitem[Parkhi et~al.(2012)Parkhi, Vedaldi, Zisserman, and Jawahar]{parkhi2012cats}
Parkhi, O.~M., Vedaldi, A., Zisserman, A., and Jawahar, C.
\newblock Cats and dogs.
\newblock In \emph{2012 IEEE conference on computer vision and pattern recognition}, pp.\  3498--3505. IEEE, 2012.

\bibitem[Perozzi et~al.(2014)Perozzi, Al-Rfou, and Skiena]{perozzi2014deepwalk}
Perozzi, B., Al-Rfou, R., and Skiena, S.
\newblock Deepwalk: Online learning of social representations.
\newblock In \emph{Proceedings of the 20th ACM SIGKDD international conference on Knowledge discovery and data mining}, pp.\  701--710, 2014.

\bibitem[Radford et~al.(2015)Radford, Metz, and Chintala]{radford2015unsupervised}
Radford, A., Metz, L., and Chintala, S.
\newblock Unsupervised representation learning with deep convolutional generative adversarial networks.
\newblock \emph{arXiv preprint arXiv:1511.06434}, 2015.

\bibitem[Razavi et~al.(2019)Razavi, Van~den Oord, and Vinyals]{razavi2019generating}
Razavi, A., Van~den Oord, A., and Vinyals, O.
\newblock Generating diverse high-fidelity images with vq-vae-2.
\newblock \emph{Advances in neural information processing systems}, 32, 2019.

\bibitem[Ribeiro et~al.(2016)Ribeiro, Singh, and Guestrin]{ribeiro2016should}
Ribeiro, M.~T., Singh, S., and Guestrin, C.
\newblock " why should i trust you?" explaining the predictions of any classifier.
\newblock In \emph{Proceedings of the 22nd ACM SIGKDD international conference on knowledge discovery and data mining}, pp.\  1135--1144, 2016.

\bibitem[Rudin(2019)]{rudin2019stop}
Rudin, C.
\newblock Stop explaining black box machine learning models for high stakes decisions and use interpretable models instead.
\newblock \emph{Nature machine intelligence}, 1\penalty0 (5):\penalty0 206--215, 2019.

\bibitem[Saul \& Roweis(2003)Saul and Roweis]{saul2003think}
Saul, L.~K. and Roweis, S.~T.
\newblock Think globally, fit locally: unsupervised learning of low dimensional manifolds.
\newblock \emph{Journal of machine learning research}, 4\penalty0 (Jun):\penalty0 119--155, 2003.

\bibitem[Sheng \& Uthus(2020)Sheng and Uthus]{sheng2020investigating}
Sheng, E. and Uthus, D.~C.
\newblock Investigating societal biases in a poetry composition system.
\newblock In \emph{Proceedings of the Second Workshop on Gender Bias in Natural Language Processing}, pp.\  93--106, 2020.

\bibitem[Sohn(2016)]{sohn2016improved}
Sohn, K.
\newblock Improved deep metric learning with multi-class n-pair loss objective.
\newblock \emph{Advances in neural information processing systems}, 29, 2016.

\bibitem[Srivastava et~al.(2014)Srivastava, Hinton, Krizhevsky, Sutskever, and Salakhutdinov]{srivastava2014dropout}
Srivastava, N., Hinton, G., Krizhevsky, A., Sutskever, I., and Salakhutdinov, R.
\newblock Dropout: a simple way to prevent neural networks from overfitting.
\newblock \emph{The journal of machine learning research}, 15\penalty0 (1):\penalty0 1929--1958, 2014.

\bibitem[Tenenbaum et~al.(2000)Tenenbaum, Silva, and Langford]{tenenbaum2000global}
Tenenbaum, J.~B., Silva, V.~d., and Langford, J.~C.
\newblock A global geometric framework for nonlinear dimensionality reduction.
\newblock \emph{science}, 290\penalty0 (5500):\penalty0 2319--2323, 2000.

\bibitem[Van Den~Oord et~al.(2017)Van Den~Oord, Vinyals, et~al.]{van2017neural}
Van Den~Oord, A., Vinyals, O., et~al.
\newblock Neural discrete representation learning.
\newblock \emph{Advances in neural information processing systems}, 30, 2017.

\bibitem[Vaswani et~al.(2017)Vaswani, Shazeer, Parmar, Uszkoreit, Jones, Gomez, Kaiser, and Polosukhin]{vaswani2017attention}
Vaswani, A., Shazeer, N., Parmar, N., Uszkoreit, J., Jones, L., Gomez, A.~N., Kaiser, L., and Polosukhin, I.
\newblock Attention is all you need.
\newblock In \emph{NIPS}, 2017.

\bibitem[Wang et~al.(2018)Wang, Singh, Michael, Hill, Levy, and Bowman]{wang2018glue}
Wang, A., Singh, A., Michael, J., Hill, F., Levy, O., and Bowman, S.~R.
\newblock Glue: A multi-task benchmark and analysis platform for natural language understanding.
\newblock In \emph{International Conference on Learning Representations}, 2018.

\bibitem[Wang \& Gupta(2015)Wang and Gupta]{wang2015unsupervised}
Wang, X. and Gupta, A.
\newblock Unsupervised learning of visual representations using videos.
\newblock In \emph{Proceedings of the IEEE international conference on computer vision}, pp.\  2794--2802, 2015.

\bibitem[Warstadt et~al.(2019)Warstadt, Singh, and Bowman]{warstadt-etal-2019-neural}
Warstadt, A., Singh, A., and Bowman, S.~R.
\newblock Neural network acceptability judgments.
\newblock \emph{Transactions of the Association for Computational Linguistics}, 7:\penalty0 625--641, 2019.
\newblock \doi{10.1162/tacl_a_00290}.
\newblock URL \url{https://aclanthology.org/Q19-1040}.

\bibitem[Wei et~al.(2019)Wei, Wu, and Cui]{wei2019deep}
Wei, X.-S., Wu, J., and Cui, Q.
\newblock Deep learning for fine-grained image analysis: A survey.
\newblock \emph{arXiv preprint arXiv:1907.03069}, 2019.

\bibitem[Weinberger \& Saul(2009)Weinberger and Saul]{weinberger2009distance}
Weinberger, K.~Q. and Saul, L.~K.
\newblock Distance metric learning for large margin nearest neighbor classification.
\newblock \emph{Journal of machine learning research}, 10\penalty0 (2), 2009.

\bibitem[Zhang et~al.(2022{\natexlab{a}})Zhang, Gu, Zhang, Bao, Chen, Wen, Wang, and Guo]{zhang2022styleswin}
Zhang, B., Gu, S., Zhang, B., Bao, J., Chen, D., Wen, F., Wang, Y., and Guo, B.
\newblock Styleswin: Transformer-based gan for high-resolution image generation.
\newblock In \emph{Proceedings of the IEEE/CVF conference on computer vision and pattern recognition}, pp.\  11304--11314, 2022{\natexlab{a}}.

\bibitem[Zhang et~al.(2022{\natexlab{b}})Zhang, Zhang, Mensah, Liu, and Mao]{zhang2022unsupervised}
Zhang, Y., Zhang, R., Mensah, S., Liu, X., and Mao, Y.
\newblock Unsupervised sentence representation via contrastive learning with mixing negatives.
\newblock In \emph{Proceedings of the AAAI Conference on Artificial Intelligence}, volume~36, pp.\  11730--11738, 2022{\natexlab{b}}.

\end{thebibliography}
\bibliographystyle{icml2024}

\newpage
\appendix
\onecolumn

\appendix
\section{Proofs}\label{apx:proofs}

\subsection{Proof of Lemma~\ref{lemma:output-space-bounds}}
\begin{manuallemma}{1}
    A G-Lipschitz function $o(\cdot)$ and a K-Lipschitz inverse function of $o(\cdot)$ returns the output space distance such that:
    \begin{equation}
        K \|\mathbf{\mathbf{v}}-\mathbf{u}\| \leq \|o(\mathbf{v})-o(\mathbf{u})\| \leq G \|\mathbf{v}-\mathbf{u}\|\,,
    \end{equation}
    where $\mathbf{\hat{v}}$ and $\mathbf{u}$ are any vector in the latent space.
\end{manuallemma}
{\it Proof.} RHS: The definition of Lipschitz continuity with constant $G$.

LHS: Starting from Lipschitz continuity with constant $K$ of the inverse function $o^{-1}(\cdot)$:
\begin{equation}
    \|o^{-1}(o(\mathbf{v})) - o^{-1}(o(\mathbf{u}))\| \leq K \|o(\mathbf{v})-o(\mathbf{u})\|\,.
\end{equation}
Using the fact that $o^{-1}(o(x)) = x$ for any $x$, we have $\|o(\mathbf{v})-o(\mathbf{u})\| \geq K \|\mathbf{v}-\mathbf{u}\|$.

\subsection{Proof of Theorem~\ref{theorem:distance-bound}}

\begin{manualtheorem}{1}
    Consider equal token importance distribution, Equation~\ref{eq:atomic-distance} returns the atomic distance such that:
    \begin{equation}
        \bar{d}_{AB} \leq C \|\mathbf{\tilde{v}}^A - \mathbf{\tilde{v}}^B\|_2\,,
    \end{equation}
    where $\mathbf{\tilde{v}}^A$ is a permutation of $\ell(\mathbf{z}^A)$ from data sample A.
\end{manualtheorem}

{\it Proof.} The key idea is using the sorted token importance vector $\delta$ and the sorting permutation matrices $\pi_A$ and $\pi_B$, such that:
\begin{equation}
    \delta := \pi_A q_A = \pi_B q_B\,,
\end{equation}
and,
\begin{equation}
\begin{split}
    \mathbf{\tilde{v}}^A & = \pi_A \mathbf{v}^A\,,\\
    \mathbf{\tilde{v}}^B & = \pi_B \mathbf{v}^B\,.
\end{split}
\end{equation}
We can rewrite Equation~\ref{eq:atomic-distance} and use H\"older's inequality:
\begin{equation}
    \bar{d} := \|\vec{\mu}_A - \vec{\mu}_B\|_1  = \|\delta \mathbf{\tilde{v}}^A - \delta \mathbf{\tilde{v}}^B\|_1 \leq \|\delta\|_2 \|\mathbf{\tilde{v}}^A - \mathbf{\tilde{v}}^B\|_2 \,.
\end{equation}

\subsection{Proof of Theorem~\ref{theorem-distance}}

\begin{manualtheorem}{2}
    
Let $c = {\sum}_{q_iq_j>0} q_iq_j$ and $c^* = {\sum}_{q_iq_j<0} q_iq_j$ for all $i\in A$ and $j\in B$, 
given data samples $A$, $B$.
Without loss of generality, $c^* = k c$ and $k\in(1,\infty)$ gives the optimal distance in Equation~\ref{eq:loss} as:
\begin{equation}
   d^* = \bigl(\frac{r_A+r_B}{2}\bigr) \frac{\sqrt{k}+1}{k-1}
\end{equation}

If $k\rightarrow 1$, $d^* \rightarrow \infty$.
\end{manualtheorem}

{\it Proof.}
We first reparameterize the distance term in Equation~\ref{eq:loss} by Equation~\ref{eq:atomic-distance}:
\begin{equation}
    \begin{split}
        \mathcal{L}_\mathcal{A} & = \sum_{i\in A,j\in B} \frac{q_i q_j}{d_{ij}}\\
        & = \sum_{q_i q_j > 0} \frac{q_i q_j}{\bar{d}_{AB}} + \sum_{q_i q_j < 0} \frac{q_i q_j}{\bar{d}_{AB} + \frac{r_A+r_B}{2}}
    \end{split}
\end{equation}

To derive an equilibrium status of the loss, we first simplify the notation by defining $d:= \bar{d}_{AB}$ and $\tilde{r} := \frac{r_A+r_B}{2}$.

\begin{equation}
    \frac{\partial \mathcal{L}_\mathcal{A}}{\partial d} = \frac{\sum_{q_iq_j> 0} -q_iq_j (d+\tilde{r})^2 + \sum_{q_iq_j< 0} -q_iq_j d^2}{d^2(d+\tilde{r})^2} = 0
\end{equation}

We further set $c \triangleq \sum_{q_iq_j> 0} q_iq_j$ and $c^* \triangleq \sum_{q_iq_j< 0} -q_iq_j$. Therefore,

\begin{equation}
    \begin{split}
        & \sum_{q_iq_j> 0} -q_iq_j (d+\tilde{r})^2 + \sum_{q_iq_j< 0} -q_iq_j d^2\\
        & = (-c + c^*) d^2 - 2c d \tilde{r} - c\tilde{r}^2\\
        & = (-c + c^*) (d - \frac{c \tilde{r}}{-c + c^*})^2  - (-c + c^*) (\frac{c \tilde{r}}{-c + c^*})^2 - c \tilde{r}^2 = 0
    \end{split}
\end{equation}

This equation can be reduced to:

\begin{equation}
    (d- \frac{c\tilde{r}}{-c + c^*})^2 = \frac{c^2 + c ( -c + c^*)}{(-c+c^*)^2} \tilde{r}^2 = \frac{c^* c}{(-c+c^*)^2} \tilde{r}^2
\end{equation}

Therefore, we can find a closed-form solution of $d$:
\begin{equation}
    d = (\frac{c}{-c+c^*} \pm \sqrt{\frac{c^* c}{(-c+c^*)^2}})\tilde{r}
\end{equation}

To this end, we can observe two unsatisfying cases from the derived formulation of $d$.
\begin{itemize}
    \item If $c = c^*$, the right hand side  divided by zero is undefined.
    \item If $c > c^*$ and recall that $d = \|\cdot\|_p \geq 0$, then $\frac{c}{c-c^*} \leq \sqrt{\frac{c^* c}{(-c+c^*)^2}}$. This results in $c^2 < c^* c$, contradict to the premise $c > c^*$.
\end{itemize}

The formulation turns out to be satisfied if and only if $c < c^*$.
A balance point exists in $d^* = \frac{c + \sqrt{c^* c}}{c^*-c} \tilde{r}$.

Suppose that $c^* = k c$ with $k > 1$, then,
\begin{equation}
    \begin{split}
        d^* & = \frac{c + \sqrt{c^* c}}{c^*-c} \tilde{r}\\
        & = \frac{c + \sqrt{k}c}{(k-1)c} \tilde{r} = \frac{\sqrt{k}+1}{k-1} \frac{r_A+r_B}{2}
    \end{split}
\end{equation}

This mathematical result indicates that optimizing $\mathcal{L}_{\mathcal{A}}$ directly depends on the relationship $k$ of the learned data sample structures.

Moreover, the balanced distance $d^*$ monotonically decreasing with respect to $k$. This statement can be proved by the partial derivative is less than zero $\forall k>1$:
\begin{equation}
    \nabla_k \frac{\sqrt{k}+1}{k-1} = -\frac{1}{2} - \sqrt{k} - \frac{1}{2}k < 0, \forall k>1
\end{equation}

\section{Discussion}\label{apx:discussion}

\paragraph{Atom modeling layer selection.}
We take $l$ as a hyperparameter to indicate that we apply atom modeling on the $l$-th hidden layer output among a total of $L$ layers.
Empirically, there is no clear pattern but the nearest layer to the raw data ($l$=1 for classifiers and $l$=L-1 for generative models) works.

\paragraph{Token importance dimension selection.}
Both in theory and empirically, selecting which dimension as token importance impacts neither training from scratch nor fine-tuning.
For simplicity, our reported experiments use the first dimension of the representation for $Q(\cdot)$.

\paragraph{Time complexity.}
The complexity of the multiplication-inverse term $E_{(A,B)\sim \mathcal{D}} \sum_{i\in A,j\in B} \frac{q_iq_j}{d_{ij}}$ is $\mathcal{O}(N^2M^2)$, where $M$ is the batch size and $N$ is the number of sub-components per data sample.
To reduce the complexity, we sample $M$ pairs of data samples in each batch and select a fixed number $S$ of sub-components from each data sample.
The used time complexity is thus $\mathcal{O}(MS)$.

When setting $S$ as the maximum of $100$ and the data size, the empirical average training time per iteration is listed in Table~\ref{tab:time-exp}.
The results show that atom modeling adds around 3-12\% training time to the standard methods and the added time depends on the used $S$ and data size.

\begin{table}[h]
    \centering
    \begin{tabular}{c|cc}
        \multirow{2}{*}{} & ImageNet-1K & MNIST \\
         & ResNet & DCGAN\\\hline
        Standard & 0.665 & 0.024 \\
        +Atom Modeling & 0.687 & 0.027 \\\hline
        Time Increased & 3.3\% & 12.5\%
    \end{tabular}
    \caption{Training time (seconds) per iteration on ImageNet-1K classsification and MNIST image synthesis.}
    \label{tab:time-exp}
\end{table}

\section{Experimental Details}
\subsection{Computation}
There are four types of computation usages in this work:
(1) The synthetic dataset experiments were using 1 RTX2070 with Max-Q with 8G capacity.
(2) The experiments of Pets, Flowers, CoLA, Poem, MNIST, CIFAR10 were using 1 Titan RTX with 24G capacity.
(3) Each experiment of ImageNet-1K was using 4 Titan RTX for two days.
(4) Each experiment of CelebA-HQ was using 6 Titan RTX for one week.

\subsection{Used Models, Baselines, and Codes}
All implementations in this work are based on PyTorch \url{https://pytorch.org/}.
(1) The synthetic dataset experiments are implemented our own.
(2) The experiments of CoLA and Poem are our implementation.
(3) The experiments of Pets and Flowers are revised from \url{https://github.com/kuangliu/pytorch-cifar} and \url{https://github.com/Skuldur/Oxford-IIIT-Pets-Pytorch}.
(4) The experiments of image generation of MNIST and CIFAR10 use exact the same script as \url{https://github.com/pytorch/examples/tree/main/dcgan}.
(5) The experiments of ImageNet-1K use exact the same script as \url{https://github.com/pytorch/examples/tree/main/imagenet}.
(6) The experiments of CelebA-HQ use exact the same script as \url{https://github.com/microsoft/StyleSwin}.

Our hyperparameter search is limited to the computation we have for this work and is as follows:
(1) The synthetic dataset experiment: We first search the best learning rate using only cross-entropy. Secondly, we use the same learning rate for all auxiliary losses and search their coefficients in \{0.5,0.2,0.1,0.05,0.02,0.01,0.005\}.
(2) The experiments of Pets, Flowers, CoLA, and Poem: We first search the best learning rate using only cross-entropy. Secondly, we use the same learning rate for all auxiliary losses and search their coefficients in \{0.05,0.02,0.01,0.005,0.002,0.001\}.
(3) The experiments of ImageNet-1K, MNIST, CIFAR10, and CelebA-HQ only use the coefficient 0.02 for atom modeling without further trials.

We summarize the baselines with their math forms or the implementation details in the following, where $h_i$ denotes the concatenation of all sub-units representations in the $i$-th data sample:
\begin{itemize}
    \item Hinge (1-norm): $\mathcal{L}_{i} = \sum_{i=1}^N \max\left\{ 0, \|h_i, h_i^+\|_1 - \|h_i, h_i^-\|_1\right\}$.
    \item Hinge (2-norm): $\mathcal{L}_{i} = \sum_{i=1}^N \max\left\{ 0, \|h_i, h_i^+\|_2 - \|h_i, h_i^-\|_2\right\}$.
    \item Rank-H~\cite{wang2015unsupervised}: $\mathcal{L}_{i} = \sum_{i=1}^N \max\left\{ 0, sim(h_i, h_i^+) - sim(h_i, h_i^-) + M \right\}, M=0.5$.
    \item SimCLR~\cite{chen2020simple}: $\mathcal{L}_{i,j} = -\log \frac{exp(sim(h_i,h_j)/\tau)}{\sum_{k=1}^{2k} \mathbbm{1}(k\neq i) exp(sim(h_i,h_k)/\tau)}$, where $(h_i, h_j)$s are positive pairs ($j$ being one data augmentation result of the $i$-th sample) and $(h_i, h_k)$s are all other possible pairs.
    \item SimCSE~\cite{gao2021simcse}: $\mathcal{L}_{i,j} = -\log \frac{exp(sim(h_i,h_j^+)/\tau)}{\sum_{j=1}^{N} exp(sim(h_i,h_j^+)/\tau)}$, where the $h^+$ is the representation produced by the same model using dropout, but at different time.
    \item MixCSE~\cite{zhang2022unsupervised}: improved SimCSE by the modified loss $\mathcal{L}_{i,j} = -\log \frac{exp(h_i^T,h_i^{'})}{C + \sum_{j=1}^{N} exp(h_i^T,SG(h_{ij}^{'}))}$, where $C=exp(h_i^Th_i^{'}/\tau) + \sum_j^N exp(h_i^Th_j^{'}), h_{ij}^{'}=\frac{\lambda h_i^{'} + (1-\lambda) h_j^{'}}{\|\lambda h_i^{'} + (1-\lambda) h_j^{'}\|_2}$.
    \item VQ+DCGAN: the quantizer~\cite{van2017neural} is implemented as \url{https://github.com/MishaLaskin/vqvae}.
    \item SSCL+DCGAN: revised SimCLR loss since no other positive pairs except for the sample itself exist in image synthesis $\mathcal{L}_{i} = -\log \frac{exp(sim(h_i,h_i)/\tau)}{\sum_{k=1}^{N} exp(sim(h_i,h_k)/\tau)}$.
\end{itemize}

\subsection{Significance Test Results}
The results of significance tests of Table~\ref{tab:results-fine-grained} are as the following Table~\ref{tab:pvalue-fine-grained}. The difference is considered significant when p-value$<0.05$ and no effect when p-value$>0.1$. 

\begin{table}[h]
    \centering
    \begin{tabular}{c|cccc}
        Method vs Cross-Entropy & Pets & Flowers & CoLA & Poem \\\hline
        Hinge (1-norm) & 0.24 & 0.15 & 0.31 & \underline{0.08} \\
        Hinge (2-norm) & \underline{0.09} & 0.19 & 0.43 & 0.14 \\
        Rank-H/SimCSE & 0.12 & \bf 0.03 & 0.13 & 0.22 \\
        SimCLR/MixCSE & 0.39 & \underline{0.09} & 0.32 & 0.24 \\
        Atom Modeling & \bf 0.03 & \bf 0.02 & \bf 0.01 & \underline{0.09} \\
    \end{tabular}
    \caption{The p-value of results comparing each method to only using cross-entropy for training.}
    \label{tab:pvalue-fine-grained}
\end{table}

\end{document}